# DQNAS: Neural Architecture Search using Reinforcement Learning

Anshumaan Chauhan
Department of Computer Science and Engineering
Florida Institute of technology
Melbourne, United States
achauhan@fit.edu

Siddhartha Bhattacharyya
Department of Computer Science and Engineering
Florida Institute of technology
Melbourne, United States
sbhattacharyya@fit.edu

S. Vadivel
Department of Computer Science
BITS Pilani, Dubai Campus
Dubai, United Arab Emirates
vadivel@dubai.bits-pilani.ac.in

*Abstract*— **Convolutional Neural Networks have been used in a variety of image related applications after their rise in popularity due to ImageNet competition. Convolutional Neural Networks have shown remarkable results in applications including face recognition, moving target detection and tracking, classification of food based on the calorie content and many more. Designing of Convolutional Neural Networks requires experts having a cross domain knowledge and it is laborious, which requires a lot of time for testing different values for different hyperparameter along with the consideration of different configurations of existing architectures. Neural Architecture Search is an automated way of generating Neural Network architectures which saves researchers from all the brute-force testing trouble, but with the drawback of consuming a lot of computational resources for a prolonged period. In this paper, we propose an automated Neural Architecture Search framework DQNAS, guided by the principles of Reinforcement Learning along with One-shot Training which aims to generate neural network architectures that show superior performance and have minimum scalability problem.**

## I. Introduction

The origin of Convolution Neural Networks (CNNs) dates to 1950, when Perceptron algorithm was invented [1]. However, Deep Learning neural networks, especially, CNNs became extremely popular after the introduction of ImageNet competition by University of Toronto in 2012. CNNs have shown promising results in the task of Image classification and Image segmentation in a variety of applications ranging from student attendance system using face recognition to medical image processing.

A CNN typically consists of the following layers: Convolutional Layers, Pooling Layers and Fully Connected Layers. The problem in designing a Neural Network involves manually selecting several parameters including number of hidden layers, number of neurons in each hidden layer, objective function to be minimized, learning rate, dropout rates, activation function to be used [2] and hyperparameters such as stride, padding and filter size in case of CNN [3]. As the range of values that can be used for these parameters is huge, there are infinite number of combinations that can be made, and this makes the manual selection quite difficult. Therefore, there are state-of-the-art architectures that are handcrafted by the experts who have expertise in the field [1], cross domain knowledge of Deep Learning, Computer Science and Optimization. However, a CNN which is performing well on one dataset, may not perform well on the other. Therefore, there is a high demand for the automated framework that takes the data as input and gives a well performing architecture as an output [4].

Designing of a Neural Network architecture can be categorized as a model selection problem [5]. Direct solution can be applying hyperparameter optimization, that is, getting the optimized values for parameters such as number of layers, activation function, etc. Hyperparameter optimization is also known as black box optimization problem because there is no explanation of the mapping done between the architecture that has been created, the performance achieved and the learning task. Three meta-modelling aspects that are used for automatic generation of CNN architecture are Hyperparameter Optimization, Evolutionary Algorithms and Reinforcement Learning.

In the past few years, there have been many fully automated algorithms and frameworks that have been developed for this task, such as:
1) MetaQNN [3]
2) BlockQNN [21]
3) GeNet [28]
4) AutoKeras [47]

and many more. Most of the research is focused on using Evolutionary Methods and Reinforcement Learning.

The paper is organized as follows: Section II contains theoretical explanation of different algorithms used for this task. Section III illustrates drawbacks faced by currently existing methodologies. Section IV briefly summarizes the working and performance of different frameworks integrated for this research and finally Section V comprised of the conclusion which concludes the paper.

## II. LITERATURE SURVEY

In this section, we will discuss about Network Architecture Search (NAS), and different Meta-modelling techniques that are used for the automatic generation of CNN architectures along with their respective drawbacks.

For classical machine learning algorithms the problem of finding optimal values for hyperparameter was resolved using techniques such as Grid Search, Random Seach, Meta Learning and Bayesian Optimization. But these methods are difficult to implement for the optimization of deep learning architecture parameters [6]. Drawback of these algorithms is that they take too much time to find the optimal values for the hyperparameters.

### A. Concept of Neural Architecture Search

Whenever we are developing a deep learning architecture for a particular application, data engineers are expected to have knowledge about what type of architecture might perform well on the given data. But, the possible number of architectures that can show good results are infinte, and hence the need for automatic architecture selection came up. The main aim behind the concept of NAS was the automation of the process of finding an architecture that show good results for a given dataset.

Zoph et al. [7] was the first one to use Reinforcement Learning to generate the values for different layers of a CNN. These different values of hyperparaters will as a whole formed the search space. They used a Controller which used to develop architetcures, which were further trained on CIFAR-10 dataset for a number of epochs. The reward function used to take the validation accuracies of last 5 epochs of the created architecture and then calculate a discounted reward which was given to Controller for training it. Finally, the algorithm used to stop either if the number of layers exceeded the maximum layer allowed or the Controller epochs have reached maimum number of iterations.

The current research area of NAS can be divided into three types as shown in Figure 1.

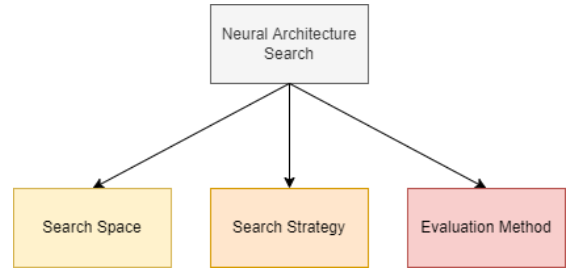

Fig. 1. Research Areas in NAS

*Search Space:* This research area focuses on what should be present in the search space. There are many methods such as Layer-by-Layer, Cell-wise and Block structures that can be used to fill up the search space. Each method has their own drawbacks, Layer-by-Layer makes the search space too large, and it takes a lot of time for the searching algorithm to find a good architecture, whereas for block and cell-based search space, they face the drawback of limited options. For example, all the search space is made using ResNet and DenseNet block structures, then there are endless possibilities that are not even being considered in this search space. It can be thought of as a trade-off between the search space and generalizability.

*Search Strategy:* Optimization algorithms that should be used to explore the search space for finding out the best possible architecture faster while maintaining a good accuracy is decided in this phase. Mostly, all the research has been broadly categorized into three parts as shown in Figure 2. More details regarding these algorithms have been provided in the following sub sections.

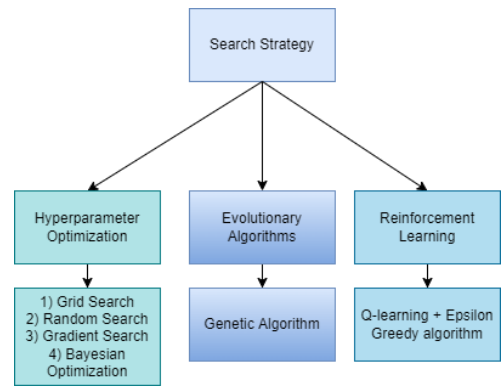

Fig. 2. Search Algorithms used to explore the Search Space of NAS

There are also two other approaches: Network Morphing and Game theory that are used, but not much research has been done in that area.

*Evaluation Method:* The way in which we should train and evaluate our models such that it takes minimum time is the main objective of this research area. There are many techniques that are proposed such as extrapolation of the accuracy curve and predicting the final accuracy, training the models for a smaller number of epochs, training the model on small dataset [8], or this is also addressed by sometimes limiting the size of Neural Network to a particular number of hidden layers. But results show that whenever we apply these strategies, there is an effect on the accuracy of the model and is also misleading when we are ranking these algorithms based on the validation accuracy. This shows that there is inverse relationship (trade-off) between performance and latency. The need for an approach that speeds up the search process without affecting the accuracy has also been mentioned in [9].

The bottleneck problem of Neural Architecture Search is the high computational time and hardware resources required by the algorithms for finding an efficient architecture. Most of the models that have shown to generate an architecture with high accuracy were susceptible to high resource consumption due to the substantial number of parameters associated with the Neural Network.

Jaafra et al. [1] briefly describes different state-of-the-art CNN architectures and different layers that composes it. A summary along with reviews of all the different meta-modelling approaches that have been applied for the task of automating the Convolutional Neural Network Architecture using Reinforcement Learning is given. Different approaches that can be used to accelerate the search process such as Early stopping, usage of distributed asynchronous framework and network transformation was mentioned in the paper.

*B. Hyperparameter Optimization*

Grid Search, Random Search Bayesian optimization-based methods were some of the algorithms that were used for Hyperparameter Optimization. Discussed earlier, we have seen the huge time complexity of Grid Search and Random Search, so researchers started working on Bayesian Optimization methods for Hyperparameter Optimization and it showed impressive results in the earlier phase of Neural Architecture Search. It works on the concept of optimizing the acquisition function and maintaining a surrogate model that learns from the previous evaluations. Surrogate models that have shown promising results are:
1) Gaussian Processes
2) Random Forests
3) Tree Structured Parzen Estimator (TSPE)

Pseudocode of the Bayesian Optimization model is shown in Figure 3.

```
Initialization :
- Place a Gaussian Process prior on f
- Observe f at n_0 points accordingly to an initial space-filling
- Set n to n_0

while n<N do:
    -Update the posterior probability distribution of f using all available data
    -Identify the maximizer x_n of the acquisition function over the valid input
     domain choices for parameters, where the acquisition function is calculated using
     the current posterior distribution
    -Observe y_n = f(x_n)
    -Increment n
end while
return point with largest f(x)
```

Fig. 3. Bayesian Optimization

Kandaswamy et al. [10] proposed an approach which used Gaussian processes for generating architectures for simple multi-layer perceptron. To find the best value from the search space, they used a distance-based optimization function, which was optimized using a transport-based algorithm. Although TSPE have shown better results than Grid Search and Random Forest, a Deep Neural Network was used by the authors for architecting a surrogate model.

Motta et al. [22] used Grid Search and Random Search for the optimization of their proposed CNN's hyperparameters that would increase its accuracy for morphological classification of mosquitoes. Their proposed methodology showed 93.5% balanced accuracy in detecting the target mosquito out of all the other insects.

Gülcü at al. [23] applied a variant of Simulated Annealing for getting optimized values of hyperparameters and searching of CNN architecture called Microcanonical Optimization Algorithm. Main advantage of using this approach was there a smaller number of parameters in the CNN architecture generated as compared to CNNs generated by other methodologies. Like [3], they limited the values for the hyperparameters used for the architecture generation. But they claim, if random values are used for the hyperparameter, then the architectures generated are quite inefficient.

Yang et al. [24] applied Particle Swarm Optimization (PSO) algorithm to fine-tune the number and size of the filters used in Convolutional and Pooling layers. They used Reinforcement Learning for the initial CNN architecture which is later fine-tuned using PSO. To have better performing architectures they have included skip connections and multi-branch as primitive operations. Additionally, replay memory algorithm was used to break the correlation between the experiences and enhancement of exploration of search space.

Li et al. [25] proposed a system to optimize the values for hyperparameters in a massively parallel fashion called ASHA. This approach targets early stopping and parallel computing as its base methodologies for tuning values of a lot of hyperparameters. Results showed that ASHA performed better than all state-of-the-art optimization algorithms and was suitable for a great extent of parallelism.

## C. Evolutionary Algorithms

Traditionally, evolutionary algorithms were used for designing the architectures for Neural Networks. Evolutionary algorithms are also known as Neuro-Evolution strategy. In [11] it was mentioned that gradient-based method can easily outperform Evolutionary Algorithms for weight optimization of Neural Networks. It was concluded that Evolutionary Algorithms should be considered only for the optimization of the architecture. There are many algorithms classified as Evolutionary algorithms [12]:
1) Genetic Algorithms
2) Evolution Strategies
3) Differential Evolutions
4) Evolution of Distribution Algorithm

However, only Genetic Algorithm have been used for this. Encoding schemes used can be broadly classified into two types: Direct and Indirect encoding. In direct encoding information such as number of neurons, connectivity between layers and activation functions are stored as genotype, whereas in indirect encoding the generation rules of the Neural Network architectures are used in the genotypes. The main advantage that led to the usage of Evolutionary algorithms was their ability to handle both continuous and discrete type of data. In NAS, discrete data can be number of layers, number of neurons, type of layer, etc. and continuous data will have hyperparameters such as Learning rate and dropout rates.

Sun et al. [4][26] designed a methodology using Genetic Algorithm, which uses ResNet and DenseNet blocks as the initial population and proposed a variable length encoding scheme that would speed up the architecture design process for these blocks. Drawback with this process was that it only uses two block structures and this in addition limits the capability of the algorithm to explore other architectural elements. Additionally, it requires a lot of computation resources for the completion.

Chen et al. [27] also proposed an Evolutionary Algorithm (EA) based meta-modelling approach, but they overcame two of the limitations of EA by focusing more on generation of Lightweight CNN architectures and using Ensemble learning on the best performing architectures. Usage of a modified squeeze fire module (inspired from SqueezeNet architecture) led to reduced number of parameters in generated architecture, and high accuracy results were achieved when tested on validation dataset. Unlike a fixed length encoding scheme used in GeNet [28], a variable length multi-level chromosome was used in this paper for encoding the CNN architecture and connectivity. In GeNet, CNN architecture was generated using a graph evolution methodology, where they linked different convolutional nodes using links and the pooling layers acted as intermediate nodes. The limitation of search space due to fixed length encoding scheme is one of the major drawbacks of the proposed approach. Another genetic algorithm-based approach EDEN, which used a fixed length encoding scheme was proposed in Dufoura et al. [29]. The chromosome was composed of two values, one was learning rate and other was the CNN architecture. On top of having a drawback of fixed size encoding scheme, pooling layers and skip connections were not incorporated in the architecture composition.

H. Cai et al. [30] designed a hardware efficient framework named AutoML, which was not only responsible for automatic design of CNN networks but also focused on model compression aspects using techniques such as Quantization and Pruning, which lets the framework have a flexible bandwidth and reduces the memory footprint. Though this methodology is now used a lot in the name of AutoKeras, it required a lot of hardware resources for computation and AutoKeras in general was used for making architecture and hyperparameter value changes in the existing state-of-the-art models. Polonskaia et al. [31] introduced an evolutionary algorithm-based NAS approach called FEDOT-NAS, which uses FEDOT framework of AutoML. It focuses on reducing the time complexity in the pipeline where training and evaluation of the network created is done. They introduced methodologies like testing on small dataset, training for a smaller number of epochs. They also compared their results with other frameworks such as AutoKeras, Auto-PyTorch, AutoCNN and CNN-GA.

## D. Reinforcement Learning

In the past few years, Reinforcement Learning algorithms are the most used methodologies for creating a best architecture from the available search space in an efficient manner. Q-learning [13] and Proximal Policy Optimization [14] processes are used for the task of exploring the search space. A simple implementation of Reinforcement Learning for Neural Architecture Search is represented in Figure 4.

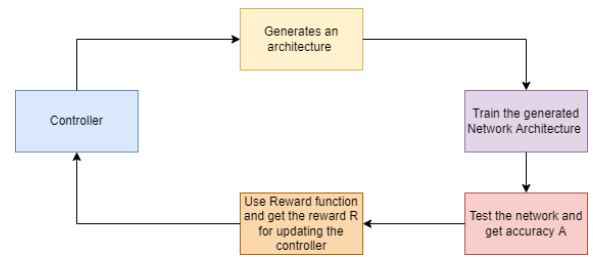

Fig. 4. NAS implementation using Reinforcement Learning

One of the main decisions one must make is to what reward function can be used that calculated a reward which guides the generator in an optimal way. For example, a reward function to guide generator in a way that it uses minimum energy while maintain a good accuracy can be:

$$R = (\alpha * Accuracy) - ((1 - \alpha) * Energy) \qquad (1)$$

Zoph et al. [7] was the first one to apply Reinforcement Learning for the task of Neural Architecture Search (NAS). Proposed methodology NASNet uses a Gradient-based approach for training the Recurrent Neural network that acts as a controller, for designing better architectures. They also have used skip connections in the search space, but irrespective of the high accuracy achieved using multi-branch and skip connections, they are computationally expensive, time expensive and requires a lot of rule-based systems for creating valid architectures (as are susceptible to compilation failure during generation of CNN).

Baker, Bowen, et al. [3] proposed a novel model MetaQNN, using Q-learning, a reinforcement learning algorithm for navigating through the search space. For efficient search they incorporated experience replay and ε-greedy algorithm, which indeed also tackles the issue of exploration versus exploitation. Validation accuracy is used to calculate the reward and updating the values in Q-table. Some constraints are put on the values that can be used for designing the architecture to reduce the search space, nevertheless the enormous number of possible combinations leads to high time complexity and huge computational resources required for running the algorithm even on a small dataset. Combined with Ensemble Learning algorithm the architectures designed using the proposed methodology outperforms the state-of-the-art architectures on Image classification datasets.

Zhong et al. [21] focused on creation of network using network blocks rather than focusing on a method where a network is designed in a layer-by-layer fashion and proposed a methodology called BlockQNN. They introduced a new encoding scheme for representation of CNNs called Network Structure Code (NSC) which consists of information related to the five parameters they considered for network generation. Using the concept of Q-learning and ε-greedy algorithm they were able to create blocks, which were stacked sequentially to make a CNN that was efficient and effective. The drawback this approach faced was that it was too data specific, it cannot be transferred to a dataset that have different input image size.

Y. Chen at al. [32] focused on creating a faster algorithm die to their limited hardware resources (1 Graphical Processing Unit available). They generated a computational cell that consisted of different layers including Convolutional Layer (can be 1x1, 1x3 or 1x5), Pooling Layer, Multi-branch connections, Skip connections and Fully Connected Layer. This cell was later stacked sequentially for a bigger CNN architecture. Gradient based methods were used for faster optimization. Instead of using a dropout layer, they used cutout [33] layer for better regularization and easier implementation.

Y. Chen et al. [34] deployed a learning agent that used epsilon greedy algorithm and experience replay for efficient navigation through the search space and generation of high performing CNN architecture. Difference from MetaQNN was usage of compact modules and stacking them sequentially rather than searching in a layer-by-layer fashion. To promote maximum flow of information between layers and strengthening of feature propagation, dense connectivity between layer was incorporated. They observed that when value of epsilon was decreased then the accuracy of CNNs were increased.

Unlike most of the research using Q-learning as their reinforcement algorithm, Mortazi et al. [35] proposed policy gradient reinforcement learning which assigns a dice index to each reward function. The average of the dice index for the last five validation accuracies of the model are averaged out and is given as input for the reinforcement learning algorithm. A new activation function names "Swish" was used instead of ReLU function. The results showed that a high accuracy was maintained by the models while maintaining a low computational cost.

Tan et al. [36] used actual latency that was observed while running model on mobiles in the optimization function rather than using FLOPs as a measure for time. A factorised hierarchical search space was used in the proposed approach MNasNet to enhance layer diversity. Proximal Policy Gradient Optimization algorithm was used to update the weights of Recurrent Neural Network (Controller).

Hsu et al. [37] proposed MONAS, a multi-objective Neural Architecture Search that uses Reinforcement Learning for generation of CNN architecture. In the paper, they proposed a mixed reward function that tried to optimize the scalability problem which will reduce the consumption of hardware resources, while maintaining a good accuracy score.

Gou et al. [38] used Inverse Reinforcement Learning, a paradigm which relies on Markov Decision Process for layer selection of CNN topology. They proposed a mirror stimuli-function, which acts as a heuristic during the designing of CNN architecture. Inverse Reinforcement Learning is used to train the mirror stimuli function, which will enhance the capability of navigating the search space efficiently without being too restricted.

Cai et al. [26] came up with a unique approach inspired from Net2Net [39] called Efficient Architecture Search (EAS), and used Network Transformations (addition, deletion or editing) on the layers of the models that have been already generated by the Bi-directional Long Short-Term Memory (Bi-LSTM). They used simple Convolutional Networks, as their base networks, and performed operations on these networks to generate an efficient network. They applied path-level network transformation in this approach, in which addition, modification and deletion of layers was performed using network transformations [40]. They also made changes in the meta-controller that was trained using Reinforcement Learning. A new tree-structured LSTM was used as a meta-controller and new operations such as Depth-separable

Convolution (DSC) and replication, were introduced in the search space.

Progressive Neural Architecture Search (PNAS), an approach given by Liu et al. [41] uses a more complex modular structure called cell, which is composed of a few blocks. A block is further composed of two out of the 8 valid operations. Like [21], the cells are then stacked sequentially to create a CNN architecture. PNAS was more than five times faster as compared to NASNet, while performing with the same level of accuracy. This was achieved by using a strategy called Sequential model-based optimization (SMBO) process, which instead of leading to a direct search in search space, performs a block-by-block search, that is, first a 1-block cell is created, which is then exploded by adding another block in that cell, and this cycle further goes on until the threshold of maximum number of blocks a cell can contain is reached.

Concept of sharing weights between the architectures created using a large computational graph was introduced in ENAS [42]. They created architectures similarly as done in PNAS, architectures were a subgraph that the larger graph and hence shared weights of previous trained networks. To train the Recurrent Neural Network they used policy gradient algorithm that focuses on maximizing the reward on the validation dataset. Due to sharing the weights and training only the best architectures on the whole validation dataset (the architectures created were initially trained only on a mini-batch dataset taken from validation dataset) this approach was proven to be a lot faster and less expensive than NASNet [7].

III. ISSUES INVOLVED IN NEURAL ARCHITECTURE SEARCH

There are many challenges involved related to the structure of a Neural Network when we are generating the architecture in an automated fashion. Setting up constraints for different hyperparameters such as what can be the maximum length of the architecture, which layer cannot be the first layer, etc. Not only does the structural issues related to Neural Network are there, the dataset used for training the intermediate models generated also plays an important role.

Apart from these, we have the problem of deciding the search strategy approaches to be used as mentioned in Section II. Issues that the existing approaches face are discussed in this section. Firstly considering we use Hyperparamter Optimization, the limitation of using Bayesian Optimization was the problem of optimizing the acquisition function due to the large number of parameters involved in Deep Neural Networks [1]. Formation of a empirical function is better as it will be optimized in a faster manner as compared to an actual optimization function [15]. Another limitation of this approach is that they can only search the models to be generated from a fixed length space. This issue of having difficulty in incorporating the connectivity of different layers in the network was disregarded when methods capable of searching in a non fixed length architecture was proposed in [16]. But there were too hard coded, in the problem of generalizability arised.

When we are using Evolutionary Algorithms as Search Strategy approach, in this case the issue is that Genetic Algorithm are suspectible to high complexity, considering different concepts such as curiosity, multi-objective optimization, etc. Some approaches have used fixed length chromosomes for the architecture representation which inhibits the Evolutionary algorithm to find the opitimal architecture, as we cannot know the length of architecture well in advance. Solving this problem by using a variable length chromosome, rises the problem of making updated mutation and crossover operations. A concept named Pareto Optimality [17] can be referred to in this search strategy. It is a concept that says we cannot enhance one objective without having an adverse affect on the other. Here tradeoff can be seen between the scalability and accuracy of the solution found [9]. Either we can find the solution quickly and deal with low accuracy or deal with scalability problem and get am optimal model with high accuracy. Further, being search bassed models, Evolutionary algorithms are quite slow, if they are not provided with some heuristics for faster convergence. In the recent researches they have started using Lightweight Convolutional Neural Network blocks such as ShuffleNet [18] and EffNet [19], and Ensemble Learning to reduce the computational complexity.

The main drawback of using Reinforcement Learning is the high time complexity. Approach proposed in [7] took 800 GPUs for 28 days to run the whole algorithm.

The drawback that the expertise is required even the automated CNN architecture design algorithm, for example, in [20] the base networks they use already are made by the experts and perform well on the given task. In [21], they have created block architectures that are purely based on Convolutional Layers and Dense Layers, exclusion of Pooling Layer has decreased the search space leading to faster generation of CNN architecture but also had a negative impact on its accuracy. It also suffers from the high computational cost problem specified above. Reinforcement Learning algorithms also suffer from the problem of exploitation versus exploration. Exploitation is the event when policy algorithm (Q-learning, Proximal Policy Gradient, etc.) choses the action which will give them the best reward repeatedly. Exploration is the event when a random action is chosen by the policy algorithm to explore the large search space. While making a CNN using a Reinforcement Learning we also must decide a rule-based constraint system which will be used to design a valid architecture from the sequence given as an output by the Controller. Constraints will involve checking whether the input image size is matching with the layer, whether the filter size is valid according to input size, etc. Summary of the drawbacks are formulated in Figure 5.

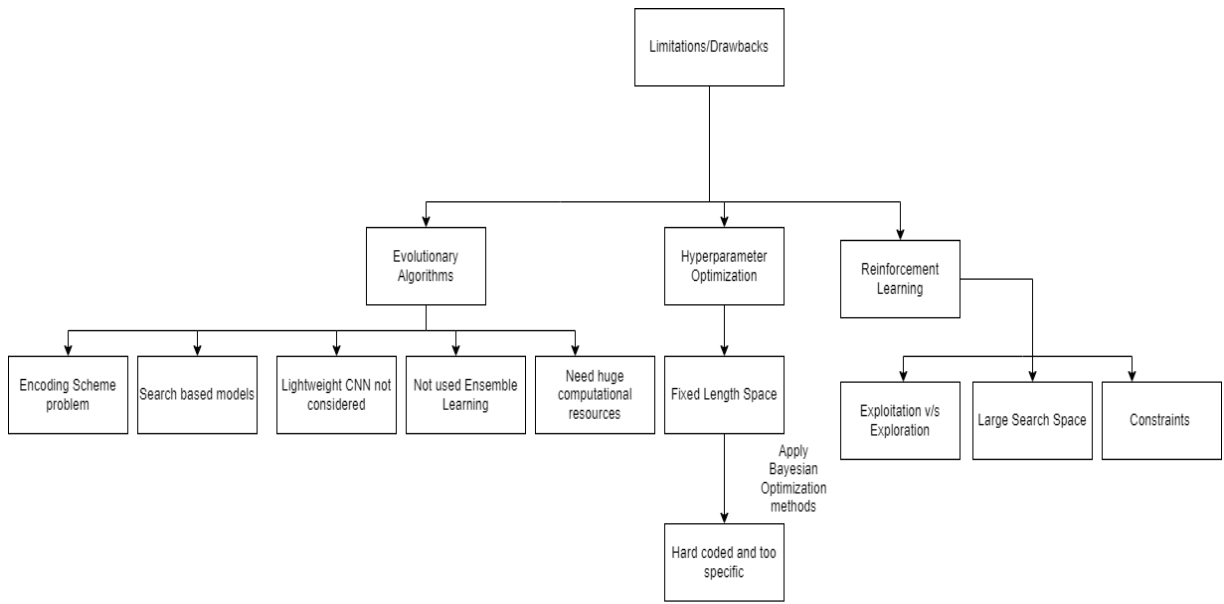

Figure 5: Drawbacks of Different Meta-Modelling techniques

## IV. DATA COLLECTION

Keras [43] is one of the most widely used library for creating different Neural Networks. Apart from the in-built functions provided for Neural Network creation, it also provides some of the commonly used datasets in Machine Learning Community such as MNIST digits classification dataset, IMDB move review dataset and many more. We are using MNIST, CIFAR10 and CIFAR100 datasets which are used for the task of training and testing a Neural Network for object detection.

MNIST, is a digit classification dataset, consisting of 60000 images for training and 10000 images for testing. It consists of total 10 classes, 1 for each integer in the range 0-9. Each image is of size 28x28 and is in a grayscale mode, therefore has only one filter layer. CIFAR10 and CIFAR 100 are object detection datasets consisting of created by University of Toronto consisting of 10 and 100 classification objects, respectively. Each of these datasets have 50000 images for training and 10000 images for testing. Images of CIFAR datasets are of size 32x32x3, as they are RGB images, therefore one filter for each color is required.

Information for the datasets is summarized in Table 1.

TABLE I. DATASETS

| Dataset | Training Images | Testing Images | Image Size | Type of Image |
|---|---|---|---|---|
| MNIST | 60000 | 10000 | 28x28x1 | Grayscale |
| CIFAR10 | 50000 | 10000 | 32x32x3 | RGB |
| CIFAR100 | 50000 | 10000 | 32x32x3 | RGB |

## V. THEORITICAL BACKGROUND ON MODEL DEVELOPMENT

### A. Double Deep Q Networks (Double DQN)

Double DQN [44] approach was proposed by a group of scientists working in Google DeepMind, to overcome the drawback faced by Q-learning algorithm of overestimating the Q-values for some actions under specific conditions. Q-learning faced the drawback of overestimating because the algorithm is trained after each step to update the Q-value based on the reward it gets for the action chosen for a given current state. Additionally, Q-learning is a very resource extensive algorithm if the number of states is large in number, as to update weights in the Q-table using Bellman Equation takes a lot of memory and time. To overcome these drawbacks Double DQN methodology was proposed, which uses a combination of two neural networks, which can be called Main and Target model, respectively. These models are trained in a fashion that they are not susceptible to the drawback of overestimating the Q-values.

In Double DQNs, Main Q network is trained after each prediction and the updated weights are calculated using a Bellman Equation. But the Target Q- network is trained only once after a few predictions (Figure 6). Rather than being trained on some data, Target Q network sets its weights to that of the Main Q network at that instant. This solves the problem of overestimating the Q-values.

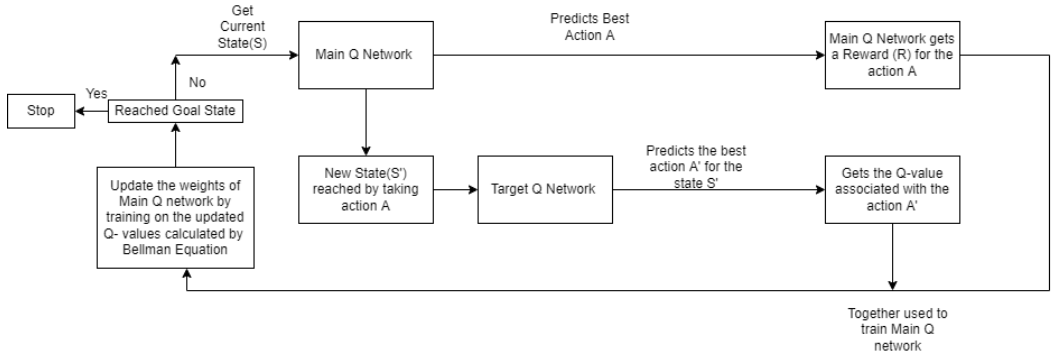

Figure 6: Double Deep Q Networks

## B. Bellman Equation

Bellman Equation is the optimality equation used for updating the Q-values for a particular state-action pair. In Double DQN, Bellman Equation calculates the updated Q-value for the Main Q network by using Reward for the action A taken by Main Q network for State S and a discounted Q value for the best associated action for the new state S.'

$$Q_M(S, A) = R + \gamma Q_T(S', A') \quad (2)$$

Where S is the current state, A is the action taken by Main Q network ($Q_M$), R is the reward achieved based on the action A, $\gamma$ is the discount factor, S' is the new state and A' is the best action associated with the state S' according to the Target Q network ($Q_T$).

## C. Epsilon Greedy Algorithm

Earlier discussed, another problem faced by Reinforcement Learning algorithm is that of Exploration vs Exploitation. To solve this issue, Epsilon Greedy algorithm is used, to make a balance between randomly exploring the search space and exploiting the best possible action. In it a random probability is generated, if it has a value greater than ε, then the action with maximum Q-value will be chosen by the Main Q network, else a random action is taken. The working of Epsilon Greedy Algorithm is shown in Figure 7.

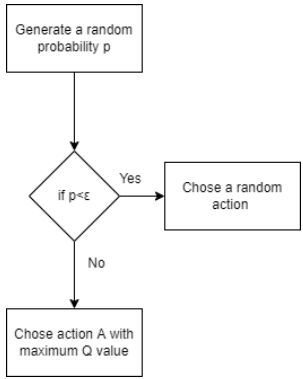

Figure 7: Working of Epsilon Greedy Algorithm

## D. Prioritized Experience Replay

Experience replay [45] is a methodology used for the training of reinforcement learning. In experience replay, a memory buffer is created, which stores all the experiences of the reinforcement learning algorithm, but they are sampled uniformly, which might contain negative experiences as well. In Prioritized Experience Replay [46], the most important transitions, are the ones sampled frequently, for an efficient training of the Reinforcement Learning algorithm. In our proposed methodology, CNN models that showed high accuracies will be taken frequently from the replay memory and used for the training of Main Q Network.

## E. One-Shot Training

One-Shot Training was basically introduced as an algorithm for knowledge transfer between the networks for the task of object recognition. The main ideology of One-Shot Training is to train a CNN with some initial knowledge rather than from scratch. Throughout the process, we will be storing the weights of CNN that are trained, later if there is a CNN that has a subpart that is like that of a previously trained model, then weights for that part will be transferred from the previously trained model to newly created model. After the training and evaluation of this new model, the weights of the sub-graph will be updated to the subgraph weights of the new model. This approach helps the model in fast evaluation and reaching its minimal error state in a shorter time interval.

Due to large combinations of layers from the search space, Neural Architecture Search faces the issue of large space complexity. Most of the existing methodologies have either reduced the search space for the algorithm or have used GPUs with high computational power, which results in not using the algorithm up to its full capacity and is not easily available to all researchers, respectively. Our contribution is in handling this gap of reducing down the time complexity without advanced hardware resources and without moderating the size of search space. In our approach, we have proposed the use of Double Deep Q- Networks, along with prioritizing experience replay and epsilon greedy algorithm, which handles the issue of learning from bad examples and exploration versus exploitation respectively. Our focus has been on the search space, by including various layers performing similar operations in a slightly different way, as they have huge effect in the model accuracy. Now once the layer is decided, the parameters associated with it are decided, for which we have variety of options available to choose from a set of values. Once the model is created and validated, we use One Shot training algorithm to transfer weights from a previously trained model if there are some common layers amongst them. By doing this, we save the time of training the model from start and are imparting some knowledge to the model beforehand.

## VI. PROPOSED METHODOLOGY

In this section we will discuss about the different components of the Proposed Methodology, including the algorithm that we have executed. This section is divided into 4 subparts: (A) Algorithm for DQNAS, (B) Search Space – description of different layers and parameters used for creation of search space, (C) Search Strategy and Action Space – model used to navigate through search space and take action (selection of layers) and (D) Evaluation metrics – metrics used for the evaluation of generated CNN.

### A. Algorithm for DQNAS

The steps performed by the algorithm for generation of an efficient and high accuracy CNN for the provided dataset is given below:

Step 1: Create the Search Space for the Reinforcement Algorithm Double DQN (Combination of Main and Target Q network). Double DQN will be using this Search Space for generation of CNN model sequences.

Step 2: Main Q Network starts generating a sequences of CNN models. For each model, it generates a random probability value p, which is then compared with the value of epsilon. If p< ε, then a random action will be taken, otherwise the action with maximum probability (Q-value) will be taken.

Step 3: Check if the chosen action satisfies all the constraints specified (Table 3), if it does then add the layer to the sequence, otherwise do a new prediction.

Step 4: Repeat Step 2, until the maximum length of the architecture is achieved.

Step 5: Repeat Step 2-4 for N number of times.

Step 6: Create CNN architectures from the sequences generated. If a model is having the error of negative dimensions while, skip the creation of the current architecture and save the model with a validation accuracy of -10.0 in the memory buffer.

Step 7: Once the model is created, compile it using each combination from learning rate value and optimizer in Table 4.

Step 8: Before training the model, use One-Shot Training for transferring the weights of some layers that were common and have the same parameter values in the previously trained models.

Step 9: After training all different versions of the model on dataset, store the model architecture and the maximum achieved validation accuracy (amongst all the versions) in memory buffer.

Step 10: Repeat Steps 6-9 for all the sequences generated by Main Q Network.

Step 11: Filter the top performing model architectures from the memory buffer and train the Main Q network using Bellman Equation.

Step 12: After every M training epoch of Main Q network, transfer the weights to Target Q network.

Step 13: Repeat steps 2-12, until N' reaches the maximum number of Controller epochs.

Figure 8 represents the various steps of the proposed algorithm.

Step 6: Create CNN architectures from the sequences generated. If a model is having the error of negative dimensions while, skip the creation of the current architecture and save the model with a validation accuracy of -10.0 in the memory buffer.

Step 7: Once the model is created, compile it using each combination from learning rate value and optimizer in Table 4.

Step 8: Before training the model, use One-Shot Training for transferring the weights of some layers that were common and have the same parameter values in the previously trained models.

Step 9: After training all different versions of the model on dataset, store the model architecture and the maximum achieved validation accuracy (amongst all the versions) in memory buffer.

Step 10: Repeat Steps 6-9 for all the sequences generated by Main Q Network.

Step 11: Filter the top performing model architectures from the memory buffer and train the Main Q network using Bellman Equation.

Step 12: After every M training epoch of Main Q network, transfer the weights to Target Q network.

Step 13: Repeat steps 2-12, until N' reaches the maximum number of Controller epochs.

Figure 8 represents the various steps of the proposed algorithm.

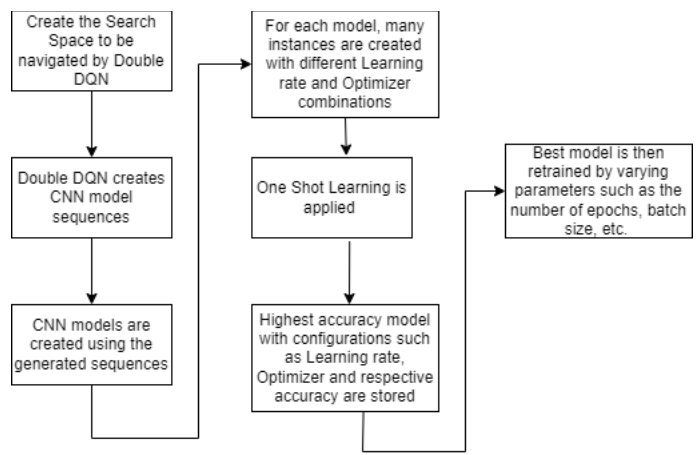

Figure 8: Proposed Methodology

## B. Search Space

Many of the previous approaches [3,51-54] have narrowed down the size of the search space to decrease the time complexity of the algorithm. But this reduced search space also directly affects the ability of the Reinforcement algorithm to sample efficient architectures. This reduction in Search Space was either in the form of limiting the type of layers that were used to create the CNN or by reducing the different parameters of a layer and their respective values.

In our proposed framework, we have focused on reducing the time complexity of the algorithm without having any reduction in the possible search space. Convolutional Neural Network usually consists of many different layers which can be divided into following 5 categories:
1) Convolutional Layer
2) Pooling Layer
3) Regularization Layer
4) Flatten Layer
5) Dense Layer

which further can have many unique layers.

In future sections, we will be discussing about the limitations that we have put on the number of layers that constitute the Q networks, those will have no effect on the Search Space that we have defined for the CNN models. This limitation on the number of layers in Q networks was done to reduce the training time. As it takes more time to train a network with many layers. Our Q network is a simple RNN constituting of few LSTM layers only. In Table 2, we have summarized different layers that are constitutes the Search Space along with the possible parameter values.

## C. Search Strategy and Action Space

In the proposed methodology, two Recurrent Neural Network (RNN) are created for the task of generating CNN sequence that will later be validated into a CNN model. In a CNN as the layer to be added further is dependent on the layers that are already added previously, therefore LSTM layers are used in the RNN model. Figure 9 represents the architecture used for the Deep Q networks. Only a few layers of RNN are used to have a faster computation that would result in less time complexity of the algorithm.

Figure 9: Deep Q network architecture

While designing a Convolutional Neural Network we have some constraints and restrictions which will lead us to create valid CNN architecture. Firstly, there are a lot of possibility that the CNN architecture generated by the algorithm leads to negative dimensions of the input image, so therefore it throws an error. To prevent the execution from stopping, we store the model architecture, and the corresponding validation accuracy is set to a negative value.

We earlier saw different parameter and their respective valid values that can be used while generating a CNN model. Now there is a possibility where we encounter an error saying that image has negative dimensions. This could be due to many reasons, such as, after some processing in the CNN, image dimensions are reduced to 5x5, and at this point it must be processed by a layer having kernel size of 7. This would be impossible, as we cannot put a 7x7 processing on a 5x5 dimensional image. For preventing our execution to stop due to this error, we skip that model's training and testing cycle, and save its architecture with a negative validation value, so that our reinforcement learning algorithm learns to avoid creating such models.

Next, we introduced few constraints [48] on the occurrence of several types of layers, such as, there should not be a Dense Layer before the occurrence of a Flatten Layer. All the constraints are tabulated in Table 3.

TABLE II. SEARCH SPACE

| Layer | Parameters | Values of the parameters |
|---|---|---|
| Convolution Layers | | |
| Conv2D | Filters (Not for DepthwiseConv2D) | $\epsilon$ {16, 32, 64, 96, 128, 160, 192, 224, 256} |
| Conv2DTranspose | Kernel Size | $\epsilon$ {3, 5, 7, 9, 11} |
| SeparableConv2D | Strides | $\epsilon$ {2, 3} (By default value is set to 1) |

| DepthwiseConv2D | | |
|---|---|---|
| | Padding | ϵ {'same', 'valid'} |
| | Kernel Initializer | ϵ {'HeNormal', 'HeUniform', 'RandomNormal', 'RandomUniform'} |
| | Bias Initializer | ϵ {'HeNormal', 'HeUniform', 'RandomNormal', 'RandomUniform'} |
| | Kernel Regularizer | ϵ {'L1', 'L2', 'L1_L2'} |
| Pooling Layers | | |
| MaxPooling2D | Pool size | ϵ {2, 3, 4, 5} |
| AveragePooling2D | Strides | ϵ {2, 3, 4, 5} |
| GlobalMaxPooling2D | Padding | ϵ {'same', 'valid'} |
| GlobalAveragePooling2D | (Not for GlobalAveragePooling2D and GlobalMaxPooling2D) | |
| Regularization Layers | | |
| Dropout | Dropout rate | ϵ {0.1, 0.2, 0.3, 0.4, 0.5, 0.6, 0.7, 0.8, 0.9} |
| BatchNormalization | (Default values for the parameters were used) | |
| Flatten Layer | | |
| Flatten | (No parameter values to be changed) | (No parameter values to be changed) |
| Dense Layer | | |
| Dense | Number of neurons | ϵ {8, 16, 32, 64, 128, 256, 512} |
| | Activation function | ϵ {'sigmoid', 'tanh', 'relu', 'elu', 'selu', 'swish'} |

Following the above-mentioned constraint once the CNN model is generated, it must be trained and tested on the dataset to know how well the model is performing. Metrics used to rate the model is validation accuracy, which is stored and is later used in the Prioritized Experience Replay to filter the better performing models from the memory buffer and train the Main Q Network on this data.

TABLE III. CONSTRAINTS FOR A VALID CNN ARCHITECTURE

| Sr. No. | Restriction |
|---|---|
| 1 | First layer of the CNN architecture should be a Convolutional layer |
| 2 | There should not be a Convolutional and Pooling layer after a Flatten Layer |
| 3 | Final Layer should always be a Dense layer having either a Softmax or a Sigmoid activation function depending on number of target classes in the dataset |
| 4 | Dropout Layer must be inserted only after a Pooling Layer (more efficient in this manner) |
| 5 | No Dense Layer can be there before a Flatten Layer |

### D. Evaluation Metrics

Training and Testing time taken by the Convolutional Neural Network has a major part in the time complexity of the whole algorithm. It takes a lot of time to train each CNN on the whole dataset for a suitable number of epochs, as it is very time consuming. Therefore, we reduced the number of training epochs for the CNN model and trained it on the whole dataset.

Each model generated is then trained with several combinations of Learning rates and Optimizers (Table 4). The values of Learning rate values are comparatively more than recommended because we are training the generated architectures in a way that we know which architecture the best is performing amongst the ones generated. Later the best architecture is trained and tested using Learning rate value of 0.01, Adam as the optimizer and 40 number of epochs. This final test accuracy will be used to compare the performance of our model with other existing methodologies.

TABLE IV. LEARNING RATE AND OPTIMIZERS

| Learning rate | ϵ {0.1, 0.2, 0.3, 0.4, 0.5, 0.6} |
|---|---|
| Optimizers | ϵ {Adam, RMSProp, SGD} |

## VII. RESULTS AND DISCUSSION

In the initial phases of the algorithm, the epsilon value is set to 1, and the decay constant has a value of 0.05. Every time when the model takes a random action epsilon value was decreased by this constant. For each of the 3 datasets, the validation dataset was 10% of the training dataset, that is, 6000 images for MNIST and 5000 images for CIFAR 10 and CIFAR100 datasets. During the model generation phase, the batch size was set to 4 and different combinations of optimizers and learning rates were used to train and test the model (Table 4). From all the possible combinations, each generated model was trained for 10 epochs, and the validation accuracy was set to the maximum of all the possible combinations generated. Our experiments were performed on CPU of Alienware Aurora R11, Intel(R) Core (TM) i9-10900F CPU @ 2.80GHz 2.81 GHz, to generate 200 models for each dataset and took around 3-4 days for the execution.

After each model is trained and validated, its model architecture, layer weights and accuracy are stored. When a new model is generated that has some common combination of layers with the previously trained models, then one shot training is used to transfer the parameter weights to this new model, this imparts some knowledge to the new generated model, and it does not start with some random initialization of weights. Table 5 contains the architecture and the accuracy of top 5 models generated for all the datasets.

Out of all these models, the top model is then trained on a learning rate value of 0.01 for 40 epochs. The top models have a huge variation in the number of learnable parameters, but they all perform with high accuracy in the respective datasets.

We have compared the accuracy obtained by DQNAS with other existing technologies and models and have tabulated the results in Table 6.

As seen in Table 6, it is evident that our proposed algorithm creates models that have comparable accuracies with the existing architectures, that were generated by scientists having cross domain knowledge of CNN, software engineering and data used. These models contain complex layers and connections and takes huge time and large computational resources for training, whereas models generated using proposed methodology are simple, takes 1 hour for training and have accuracy difference of mere 5% as compared to complex deep learning CNN architectures.

TABLE V. TOP 5 ARCHITECTURES CREATED FOR EACH DATASET

| | MNIST dataset | |
|---|---|---|
| 1 | [('conv2dtranspose', 160, 7, 2, 'valid', 'HeNormal', 'RandomUniform', 'l1_l2'), ('conv2dtranspose', 128, 9, 2, 'valid', 'RandomNormal', 'RandomNormal', 'l1'), ('separableconv2d', 96, 5, 3, 'same', 'HeNormal', 'RandomNormal', 'l1'), ('conv2dtranspose', 192, 7, 3, 'same', 'HeUniform', 'HeUniform', 'l1'), ('conv2d', 96, 5, 3, 'valid', 'HeUniform', 'RandomUniform', 'l2'), ('separableconv2d', 128, 9, 2, 'same', 'RandomUniform', 'RandomUniform', 'l1'), 'Flatten', (10, 'softmax')] | 84.76% |
| 2 | [('conv2dtranspose', 16, 3, 3, 'valid', 'RandomNormal', 'HeNormal', 'l2'), ('conv2d', 224, 9, 2, 'same', 'RandomNormal', 'RandomUniform', 'l2'), ('conv2dtranspose', 256, 5, 2, 'same', 'HeNormal', 'RandomNormal', 'l1'), ('separableconv2d', 192, 5, 2, 'same', 'HeNormal', 'HeUniform', 'l1_l2'), ('conv2d', 128, 5, 3, 'valid', 'RandomUniform', 'HeNormal', 'l1'), ('conv2dtranspose', 128, 5, 3, 'valid', 'HeUniform', 'HeUniform', 'l1_l2'), 'Flatten', (10, 'softmax')] | 82.65% |
| 3 | [('conv2d', 224, 3, 3, 'valid', 'HeUniform', 'HeUniform', 'l1_l2'), ('conv2dtranspose', 128, 9, 3, 'valid', 'RandomNormal', 'RandomNormal', 'l1_l2'), ('conv2dtranspose', 160, 7, 3, 'valid', 'HeNormal', 'RandomUniform', 'l1'), ('conv2d', 32, 7, 2, 'same', 'RandomNormal', 'RandomUniform', 'l2'), ('depthwiseconv2d', 7, 3, 'valid', 'HeUniform', 'HeUniform', 'l1_l2'), ('conv2dtranspose', 32, 3, 3, 'same', 'HeNormal', 'HeNormal', 'l1_l2'), 'Flatten', (10, 'softmax')] | 81.97% |
| 4 | [('conv2dtranspose', 192, 3, 3, 'same', 'HeNormal', 'RandomUniform', 'l1_l2'), ('depthwiseconv2d', 3, 2, 'same', 'RandomNormal', 'HeNormal', 'l1'), ('conv2d', 192, 5, 2, 'same', 'HeUniform', 'HeUniform', 'l1_l2'), ('conv2d', 192, 7, 2, 'same', 'HeUniform', 'HeNormal', 'l1_l2'), ('depthwiseconv2d', 9, 2, 'same', 'RandomNormal', 'RandomNormal', 'l2'), ('conv2d', 96, 3, 3, 'valid', 'RandomUniform', 'HeUniform', 'l1'), 'Flatten', (10, 'softmax')] | 81.82% |
| 5 | [('conv2dtranspose', 128, 7, 3, 'same', 'RandomUniform', 'HeUniform', 'l1'), ('separableconv2d', 224, 9, 3, | 81.04% |
| | 'same', 'HeUniform', 'HeUniform', 'l2'), ('conv2dtranspose', 96, 5, 3, 'same', 'HeUniform', 'HeNormal', 'l1'), ('conv2d', 192, 9, 3, 'valid', 'HeUniform', 'RandomNormal', 'l1'), ('conv2d', 128, 5, 3, 'same', 'HeNormal', 'HeUniform', 'l1_l2'), ('conv2dtranspose', 160, 7, 3, 'same', 'HeUniform', 'HeNormal', 'l1'), 'Flatten', (10, 'softmax')] | |
|  | CIFAR 10 dataset | |
| 1 | [('conv2dtranspose', 160, 5, 2, 'valid', 'RandomUniform', 'RandomUniform', 'l2'), ('depthwiseconv2d', 9, 2, 'same', 'RandomNormal', 'HeUniform', 'l1'), ('maxpool2d', 5, 2, 'valid') , ('dropout', 0.3), ('conv2d', 192, 7, 2, 'same', 'RandomUniform', 'HeUniform', 'l1'), ('globalavgpool2d', 'valid'), 'Flatten', (10, 'softmax')] | 63.05 |
| 2 | [('conv2dtranspose', 160, 9, 3, 'same', 'RandomNormal', 'RandomNormal', 'l2'), ('conv2dtranspose', 192, 3, 3, 'valid', 'RandomNormal', 'HeNormal', 'l2'), ('avgpool2d', 7, 2, 'same'), ('conv2dtranspose', 224, 3, 3, 'same', 'RandomNormal', 'HeUniform', 'l2'), ('depthwiseconv2d', 9, 3, 'same', 'RandomUniform', 'RandomUniform', 'l1_l2'), , 'Flatten', (128, 'elu'), (10, 'softmax')] | 61.21 |
| 3 | [('conv2d', 128, 7, 2, 'same', 'RandomUniform', 'HeNormal', 'l2'), ('depthwiseconv2d', 9, 2, 'valid', 'RandomNormal', 'HeNormal', 'l2'), ('conv2d', 256, 3, 2, 'same', 'HeNormal', 'HeNormal', 'l1'), ('maxpool2d', 9, 2, 'valid'), ('conv2dtranspose', 192, 7, 3, 'valid', 'RandomUniform', 'HeNormal', 'l1_l2'), ('avgpool2d', 9, 2, 'valid'), 'Flatten', (10, 'softmax')] | 60.79 |
| 4 | [('conv2d', 192, 7, 3, 'same', 'RandomNormal', 'HeUniform', 'l1_l2'), ('maxpool2d', 7, 2, 'valid'), ('maxpool2d', 9, 2, 'same'), ('conv2d', 16, 9, 3, 'same', 'RandomUniform', 'HeUniform', 'l1'), ('conv2d', 192, 5, 2, 'same', 'HeUniform', 'RandomNormal', 'l1_l2'), 'Flatten', (256, 'tanh'), (10, 'softmax')] | 60.41 |
| 5 | [('conv2d', 16, 3, 2, 'valid', 'HeNormal', 'RandomNormal', 'l2'), ('depthwiseconv2d', 5, 2, 'same', 'RandomNormal', 'HeNormal', 'l1_l2'), ('conv2dtranspose', 224, 7, 2, 'same', 'HeUniform', 'HeUniform', 'l1'), ('globalavgpool2d', 'same'), ('dropout', 0.5), ('conv2dtranspose', 160, 3, 3, 'valid', 'RandomNormal', 'RandomNormal', 'l1_l2'), 'Flatten', (10, 'softmax')] | 60.02 |
| colspan | CIFAR 100 dataset | |
| 1 | [('conv2d', 64, 9, 2, 'same', 'RandomNormal', 'HeNormal', 'l2'), ('avgpool2d', 3, 3, 'valid'), ('conv2dtranspose', 256, 3, 3, 'valid', 'RandomNormal', 'RandomUniform', 'l1_l2'), ('maxpool2d', 3, 3, 'valid'), 'BatchNormalization', 'Flatten', (256, 'swish'), (100, 'softmax')] | 38.98 |
| 2 | [('conv2d', 96, 3, 3, 'valid', 'HeNormal', 'RandomNormal', 'l2'), ('conv2d', 192, 7, 2, 'same', 'RandomUniform', 'HeUniform', 'l1'), ('maxpool2d', 3, 3, 'valid'), ('conv2dtranspose', 256, 9, 3, 'same', 'HeUniform', 'RandomNormal', 'l1'), ('conv2d', 16, 9, 3, 'same', 'HeNormal', 'HeUniform', 'l1'), ('maxpool2d', 7, 3, 'same'), 'Flatten', (100, 'softmax')] | 29.04 |
| 3 | [('conv2d', 160, 9, 2, 'valid', 'RandomNormal', 'HeUniform', 'l2'), ('globalavgpool2d', 'same'), ('depthwiseconv2d', 9, 2, 'valid', 'HeNormal', 'RandomNormal', 'l2'), ('depthwiseconv2d', 5, 3, 'valid', 'HeUniform', 'RandomNormal', 'l1_l2'), ('conv2dtranspose', 224, 7, 2, 'same', 'RandomUniform', 'RandomUniform', 'l2'), ('maxpool2d', 3, 3, 'valid'), 'Flatten', (100, 'softmax')] | 26.93 |
| 4 | [('conv2dtranspose', 64, 3, 2, 'valid', 'RandomNormal', 'HeUniform', 'l1_l2'), ('depthwiseconv2d', 3, 3, 'same', 'HeUniform', 'RandomUniform', 'l1'), ('separableconv2d', 192, 7, 2, 'same', 'RandomUniform', 'HeNormal', 'l1_l2'), ('maxpool2d', 5, 3, 'same'), ('dropout', 0.2), 'BatchNormalization', 'Flatten', (100, 'softmax')] | 25.56 |
| 5 | [('conv2d', 224, 7, 3, 'same', 'RandomUniform', 'RandomUniform', 'l1_l2'), ('depthwiseconv2d', 3, 2, 'valid', 'HeUniform', 'HeNormal', 'l1'), ('conv2dtranspose', 16, 5, 3, 'valid', 'HeNormal', 'RandomNormal', 'l2'), ('conv2dtranspose', 32, 9, 3, 'valid', 'HeUniform', 'RandomNormal', 'l1_l2'), ('depthwiseconv2d', 7, 3, 'valid', 'RandomNormal', 'RandomNormal', 'l2'), 'Flatten', (128, 'relu'), (100, 'softmax')] | 25.52 |

TABLE VI. ACCRUACY PERCENTAGE COMPARISON WITH EXISTING STATE-OF-THE-ART ARCHITECTURES AND METHODOLOGIES

| Models/Methodologies | MNIST | CIFAR10 | CIFAR100 |
|---|---|---|---|
| MetaQNN (12 layers) | 99.56 | 92.68 | 72.86 |
| VGGNet (16 layers) | - | 92.75 | - |
| FitNet (19 layers) | 99.49 | 91.61 | 64.96 |
| Proposed Methodology (8 layers) | 97.91 | 88.07 | 69.15 |

## VIII. CONCLUSION

Use of Neural Networks has been increased exponentially in the past few years, but it also generates the problem of creating domain specific high performance neural network architecture. There are a lot of meta-modelling approaches used for automatic generation of neural network architecture, but either they have a requirement of training the algorithm on the dataset using a lot of GPUs on several days or they restrict the search space make force the algorithm to follow a pattern. Our DQNAS approach, solves the problem of time complexity as well as the resource requirements. Moreover, the top architectures created can also be tested and used on other related datasets. While the DQNAS is a simple algorithm using reinforcement learning to generate CNN architectures, users can easily modify the constraints specified to create a sequence that imitates other well performing architecture.

In the future work, we would like to minimize the time complexity by incorporating the concept of predicted accuracies, which would eliminate all those architectures that the model predicts will perform poorly on the given dataset. We will incorporate other techniques for fastened training of the CNN architecture, such as usage of REINFORCE gradient for updating weights and trying out the effect of reducing the training dataset on model training. There might be instances where the total number of learnable parameters in the CNN architecture are huge, and the system may fail to create such model. To handle such cases, we will put a threshold, and if the parameters are exceeding that number, then that model should be skipped.

## REFERENCES

[1] Jaafra, Yesmina, et al. "A review of meta-reinforcement learning for deep neural networks architecture search." arXiv preprint arXiv:1812.07995 (2018).

[2] Chauhan, Anshumaan, et al. "LPRNet: A Novel Approach for Novelty Detection in Networking Packets." International Journal of Advanced Computer Science and Applications 13.2 (2022).

[3] Baker, Bowen, et al. "Designing neural network architectures using reinforcement learning." arXiv preprint arXiv:1611.02167 (2016).

[4] Sun, Yanan, et al. "Automatically evolving cnn architectures based on blocks." arXiv preprint arXiv:1810.11875 (2018).

[5] Suganuma, Masanori, Shinichi Shirakawa, and Tomoharu Nagao. "A genetic programming approach to designing convolutional neural network architectures." Proceedings of the genetic and evolutionary computation conference. 2017.

[6] Liashchynskyi, Petro, and Pavlo Liashchynskyi. "Grid search, random search, genetic algorithm: a big comparison for NAS." arXiv preprint arXiv:1912.06059 (2019).

[7] Zoph, Barret, and Quoc V. Le. "Neural architecture search with reinforcement learning." arXiv preprint arXiv:1611.01578 (2016).

[8] Real, Esteban, et al. "Regularized evolution for image classifier architecture search." Proceedings of the aaai conference on artificial intelligence. Vol. 33. No. 01. 2019.

[9] Elsken, Thomas, Jan Hendrik Metzen, and Frank Hutter. "Neural architecture search: A survey." The Journal of Machine Learning Research 20.1 (2019): 1997-2017.

[10] Kandasamy, Kirthevasan, et al. "Neural architecture search with bayesian optimisation and optimal transport." Advances in neural information processing systems 31 (2018).

[11] Floreano, Dario, Peter Dürr, and Claudio Mattiussi. "Neuroevolution: from architectures to learning." Evolutionary intelligence 1.1 (2008): 47-62.

[12] Corne, David, and Michael A. Lones. "Evolutionary algorithms." Handbook of Heuristics. Springer, Cham, 2018. 409-430.

[13] Singh, Satinder P., and Richard C. Yee. "An upper bound on the loss from approximate optimal-value functions." Machine Learning 16.3 (1994): 227-233.

[14] Byun, Ju-Seung, Byungmoon Kim, and Huamin Wang. "Proximal Policy Gradient: PPO with Policy Gradient." arXiv preprint arXiv:2010.09933 (2020).

[15] Klein, Aaron, et al. "Fast bayesian optimization of machine learning hyperparameters on large datasets." Artificial intelligence and statistics. PMLR, 2017.

[16] Bergstra, James, Daniel Yamins, and David Cox. "Making a science of model search: Hyperparameter optimization in hundreds of dimensions for vision architectures." International conference on machine learning. PMLR, 2013.

[17] Luc, Dinh The. "Pareto optimality." Pareto optimality, game theory and equilibria (2008): 481-515.

[18] Ma, Ningning, et al. "Shufflenet v2: Practical guidelines for efficient cnn architecture design." Proceedings of the European conference on computer vision (ECCV). 2018.

[19] Freeman, Ido, Lutz Roese-Koerner, and Anton Kummert. "Effnet: An efficient structure for convolutional neural networks." 2018 25th ieee international conference on image processing (icip). IEEE, 2018.

[20] Cai, Han, et al. "Efficient architecture search by network transformation." Proceedings of the AAAI Conference on Artificial Intelligence. Vol. 32. No. 1. 2018.

[21] Zhong, Zhao, et al. "Practical block-wise neural network architecture generation." Proceedings of the IEEE conference on computer vision and pattern recognition. 2018.

[22] Motta, Daniel, et al. "Optimization of convolutional neural network hyperparameters for automatic classification of adult mosquitoes." Plos one 15.7 (2020): e0234959.

[23] Gülcü, Ayla, and Zeki Kuş. "Hyper-parameter selection in convolutional neural networks using microcanonical optimization algorithm." IEEE Access 8 (2020): 52528-52540.

[24] Yang, Yuxuan, et al. "A CNN identified by reinforcement learning-based optimization framework for EEG-based state evaluation." Journal of Neural Engineering 18.4 (2021): 046059.

[25] Li, Liam, et al. "Massively parallel hyperparameter tuning." arXiv preprint arXiv:1810.05934 5 (2018).

[26] Sun, Yanan, et al. "Completely automated CNN architecture design based on blocks." IEEE transactions on neural networks and learning systems 31.4 (2019): 1242-1254.

[27] Chen, Yushi, et al. "Automatic design of convolutional neural network for hyperspectral image classification." IEEE Transactions on Geoscience and Remote Sensing 57.9 (2019): 7048-7066.

[28] Xie, Lingxi, and Alan Yuille. "Genetic cnn." Proceedings of the IEEE international conference on computer vision. 2017.

[29] Dufourq, Emmanuel, and Bruce A. Bassett. "Eden: Evolutionary deep networks for efficient machine learning." 2017 Pattern Recognition Association of South Africa and Robotics and Mechatronics (PRASA-RobMech). IEEE, 2017.

[30] Cai, Han, et al. "Automl for architecting efficient and specialized neural networks." IEEE Micro 40.1 (2019): 75-82.

[31] Polonskaia, Iana S., Ilya R. Aliev, and Nikolay O. Nikitin. "Automated evolutionary design of CNN classifiers for object recognition on satellite images." Procedia Computer Science 193 (2021): 210-219.

[32] Chen, Yushi, et al. "Automatic design of convolutional neural network for hyperspectral image classification." IEEE Transactions on Geoscience and Remote Sensing 57.9 (2019): 7048-7066.

[33] DeVries, Terrance, and Graham W. Taylor. "Improved regularization of convolutional neural networks with cutout." arXiv preprint arXiv:1708.04552 (2017).
[34] Chen, Yifang, et al. "Automated design of neural network architectures with reinforcement learning for detection of global manipulations." IEEE Journal of Selected Topics in Signal Processing 14.5 (2020): 997-1011.
[35] Mortazi, Aliasghar, and Ulas Bagci. "Automatically designing CNN architectures for medical image segmentation." International Workshop on Machine Learning in Medical Imaging. Springer, Cham, 2018.
[36] Tan, Mingxing. "MnasNet: Towards Automating the Design of Mobile Machine Learning Models." (2018).
[37] Hsu, Chi-Hung, et al. "Monas: Multi-objective neural architecture search using reinforcement learning." arXiv preprint arXiv:1806.10332 (2018).
[38] Guo, Minghao, et al. "Irlas: Inverse reinforcement learning for architecture search." Proceedings of the IEEE/CVF Conference on Computer Vision and Pattern Recognition. 2019.
[39] Chen, Tianqi, Ian Goodfellow, and Jonathon Shlens. "Net2net: Accelerating learning via knowledge transfer." arXiv preprint arXiv:1511.05641 (2015).
[40] Cai, Han, et al. "Path-level network transformation for efficient architecture search." International Conference on Machine Learning. PMLR, 2018.
[41] Liu, Chenxi, et al. "Progressive neural architecture search." Proceedings of the European conference on computer vision (ECCV). 2018.
[42] Pham, Hieu, et al. "Efficient neural architecture search via parameters sharing." International conference on machine learning. PMLR, 2018.
[43] Chollet, François. "keras." (2015).
[44] Mnih, Volodymyr, et al. "Asynchronous methods for deep reinforcement learning." International conference on machine learning. PMLR, 2016.
[45] Zhang, Shangtong, and Richard S. Sutton. "A deeper look at experience replay." arXiv preprint arXiv:1712.01275 (2017).
[46] Schaul, Tom, et al. "Prioritized experience replay." arXiv preprint arXiv:1511.05952 (2015).
[47] Jin, Haifeng, Qingquan Song, and Xia Hu. "Auto-keras: An efficient neural architecture search system." Proceedings of the 25th ACM SIGKDD international conference on knowledge discovery & data mining. 2019.
[48] Alzubaidi, Laith, et al. "Review of deep learning: Concepts, CNN architectures, challenges, applications, future directions." Journal of big Data 8.1 (2021): 1-74.